\pdfoutput=1

\documentclass[11pt]{article}
\usepackage{emnlp2021}

\usepackage{times}
\usepackage{latexsym}
\usepackage[T1]{fontenc}
\usepackage[utf8]{inputenc}
\usepackage{multirow}
\usepackage{booktabs}
\usepackage{xspace}
\usepackage{microtype}
\usepackage{enumitem}
\usepackage{balance}
\usepackage{tikz}
\usepackage{subcaption}
\usepackage[hang,flushmargin]{footmisc} % to remove indents in footnote

\usetikzlibrary{pgfplots.groupplots}

\newcommand\tydi{T{\small Y}D{\small I}\xspace}
\newcommand\xortydi{X{\small OR}-T{\small Y}D{\small I}\xspace}
\newcommand\mrtydi{Mr.~T{\small Y}D{\small I}\xspace}

\newcommand{\ignore}[1]{}

\title{\mrtydi: A Multi-lingual Benchmark for Dense Retrieval}

\author{Xinyu Zhang, Xueguang Ma, Peng Shi, and Jimmy Lin \\ [1ex]
  David R. Cheriton School of Computer Science \\
  University of Waterloo
}

\begin{document}
\maketitle
\begin{abstract}
We present \mrtydi, a multi-lingual benchmark dataset for mono-lingual retrieval in eleven typologically diverse languages, designed to evaluate ranking with learned dense representations.
The goal of this resource is to spur research in dense retrieval techniques in non-English languages, motivated by recent observations that existing techniques for representation learning perform poorly when applied to out-of-distribution data.
As a starting point, we provide zero-shot baselines for this new dataset based on a multi-lingual adaptation of DPR that we call ``mDPR''.
Experiments show that although the effectiveness of mDPR is much lower than BM25, dense representations nevertheless appear to provide valuable relevance signals, improving BM25 results in sparse--dense hybrids.
In addition to analyses of our results, we also discuss future challenges and present a research agenda in multi-lingual dense retrieval.
\mrtydi can be downloaded at \url{https://github.com/castorini/mr.tydi}.
\end{abstract}

\section{Introduction}

Retrieval approaches based on learned dense representations, typically derived from transformers, form an exciting new research direction that has received much attention of late.
These dense retrieval techniques generally adopt a supervised approach to representation learning, where a labeled dataset is used to train two encoders---one for the queries and the other for texts from the corpus to be retrieved---whose output representation vectors are then compared with a simple comparison function such as inner product.
Retrieval against a large text corpus is typically formulated as nearest neighbor search and efficiently executed using off-the-shelf libraries.
In the literature, this is known as a ``bi-encoder'' design.
Well-known examples include DPR~\cite{karpukhin-etal-2020-dense}, ANCE~\cite{xiong-etal-2021-ance-iclr}, and ColBERT~\cite{Khattab_Zaharia_SIGIR2020}, but there is much recent work along these lines~\cite{Gao_etal_arXiv2020_CLEAR,Hofstatter:2010.02666:2020,Hofstatter_etal_SIGIR2021,lin-etal-2021-batch}, just to list a few papers.

Like all methods based on supervised machine learning, the effectiveness of the trained models on ``out of distribution'' (OOD) samples is an important issue since it concerns model robustness and generalizability.
For dense retrieval, training data typically comprise (query, relevant passage) pairs, and in this context, OOD could mean that (1) the passage encoder is fed text from a different domain, genre, register, etc. than the training data, (2) the query encoder is fed queries that are different from the training queries, (3) the relationship between the inputs at inference time is different from the training samples (e.g., task variations), or (4) a combination of all of the above.

\begin{figure*}[t]
\centering
\includegraphics[width=\textwidth]{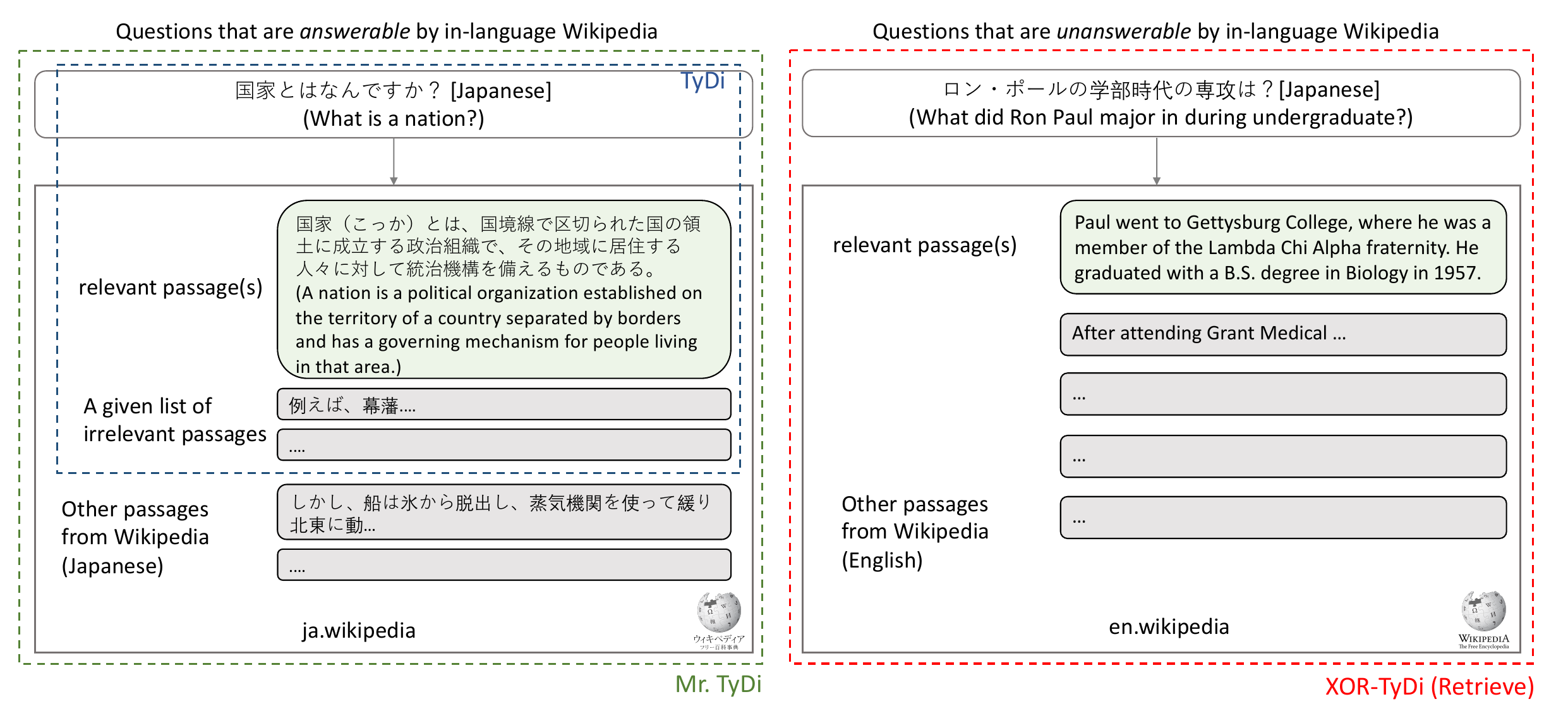}
\caption{Comparison between \tydi, \xortydi, and \mrtydi with an example in Telugu. The green blocks indicate relevant passages and the red blocks indicate non-relevant passages.}
\label{fig:dataset-cmp}
\end{figure*}

It is, in fact, already known that dense retrieval techniques generalize poorly across different corpora, queries, tasks, etc.
Recently, \citet{Thakur:2104.08663:2021} constructed a benchmark to  specifically evaluate the zero-shot transfer capabilities of dense retrieval models by creating a framework that unifies over a dozen retrieval datasets spanning diverse domains.
In a zero-shot setting, the authors found that BM25 remained the most effective overall.
That is, dense retrieval techniques trained on one dataset can spectacularly fail on another dataset---exactly the out-of-distribution challenges we discussed above.
In contrast, BM25, ``just works'' regardless of the corpus and queries, even though on ``in distribution'' samples, dense retrieval models are unequivocally more effective.
Thus, learned dense representations are not as general and robust as BM25 representations.

This paper focuses on another aspect of the generalizability and robustness of learned representations for ranking:\
What if the encoders are applied to texts in languages different from the one they are trained in?
Our focus is on {\it mono-lingual} retrieval in non-English languages (e.g., Bengali queries against Bengali documents) rather than {\it cross-lingual} retrieval, where documents and queries are in different languages (e.g., English queries against Arabic documents).

We view this work as having three main contributions:\
First, we construct and share \mrtydi, a multi-lingual benchmark dataset for mono-lingual retrieval in eleven diverse languages, designed to evaluate ranking with learned dense representations.
This dataset can be viewed as the ``open-retrieval'' condition of the \tydi multi-lingual question answering (QA) dataset~\cite{clark-etal-2020-tydi}, and ``Mr'' in \mrtydi stands for ``multi-lingual retrieval''.
We describe the construction of this dataset and how it is different from existing resources.
Second, we report zero-shot baselines for \mrtydi, including a dense retrieval method based on a multi-lingual version of DPR~\cite{karpukhin-etal-2020-dense} that we call ``mDPR''.
Third, we present a number of initial findings about baseline results that highlight future challenges and begin to define a research agenda in multi-lingual dense retrieval.
Most interestingly, we find that although the zero-shot effectiveness of mDPR is much worse than BM25, dense representations appear to provide valuable relevance signals, improving BM25 results in sparse--dense hybrids.

\section{Background and Related Work}

In presenting a new benchmark dataset, one important question to answer is:\ Why is a new resource needed?
We begin by addressing this question.
The introduction already lays out the intellectual motivation for our work.
Thus, we focus here on explaining why existing datasets are not sufficient.
The answer is summarized in Figure~\ref{fig:dataset-cmp}.

\mrtydi is constructed from \tydi~\cite{clark-etal-2020-tydi}, a question answering dataset covering eleven typologically diverse languages.
For each language, the creators provided annotators with a prompt (the first 100 characters of a Wikipedia article), who were asked to write a question that cannot answered by the snippet.
Then, for each question, annotators were given the top Wikipedia article returned by Google search and asked to label the relevance of each passage in the article as well as to identify a minimal answer span (if possible).
Given this procedure, the answer to the question may or may not be found in the passages from the selected article.
The answer passages are always in the same language as the questions.
Note that the questions in different languages are not comparable as they are created independently rather than through translation.

The weakness of \tydi from our perspective is that it is essentially a machine reading comprehension task like SQuAD~\cite{rajpurkar-etal-2016-squad} because candidate passages are all included as part of the dataset (i.e., the passages are from the top Wikipedia article returned by Google search).
Instead, we need a resource akin to what QA researchers call the ``open-domain'' or ``open-retrieval'' task, where the problem involves retrieval from a much larger corpus (e.g., all of Wikipedia)~\cite{chen-etal-2017-reading}.
Thus, at a high level, \mrtydi can be viewed as an open-retrieval extension to \tydi.

\citet{Asai2021XORQC} created \xortydi, a cross-lingual QA dataset built on \tydi by annotating answers in English Wikipedia for questions \tydi considered unanswerable in the original source (non-English) language.
This was accomplished by randomly sampling 5,000 unanswerable (non-English) questions from \tydi, and then searching English Wikipedia articles for answers.
Specifically, each non-English question was first translated into English; then, annotators were given the top-ranked English Wikipedia articles and asked to label passages containing the answer.

The \xortydi dataset contains three overlapping tasks, but all of them are focused on the cross-lingual condition.
Among the three tasks (\textsc{Xor-EnglishSpan}, \textsc{Xor-Retrieve}, and \textsc{Xor-Full}), \textsc{XOR-Retrieve} is most comparable to our work, but the retrieval target for the task is explicitly English Wikipedia articles rather than Wikipedia articles in the question's language.
While the \textsc{Xor-Full} task requires systems to select answer spans from both English and target-language Wikipedia articles, the dataset does not provide ground truth for the intermediate retrieval step, thus it cannot be used for our evaluations.

The annotations of \xortydi do not allow us to examine mono-lingual retrieval in non-English languages because the creators started with ``unanswerable'' non-English \tydi{} questions.
Furthermore, since all answer passage annotations were performed on English Wikipedia, this doesn't help if we are interested in, for example, searching Finnish Wikipedia with Finnish questions.

Another point of comparison worth discussing is MKQA~\cite{Longpre:2007.15207:2020}, which comprises 10k question--answer pairs aligned across 26 typologically diverse languages.
Questions are paired with exact answers in the different languages, and evaluation is conducted in the open-retrieval setting by matching those answers in retrieved text.
There are two main differences between our work and MKQA:
First, MKQA was created via translation in order to achieve cross-lingual alignment.
In addition to the possible translation artifacts that such a process might introduce, which~\citet{clark-etal-2020-tydi} discussed at length, we argue that forced alignment creates non-natural questions, for the simple reason that speakers of different languages are likely to be interested in different topics.
This is different from the ``geographically dependent'' questions that MKQA tries to avoid.
Take the question ``{\it who starred in the movie bridge over the river kwai}'' as an example:
While it does not involve any geographical preference, the question is probably less likely to be asked in, say, Swahili, compared to in English or Thai.
Second, the builders of MKQA explicitly made the decision to create ``retrieval-independent answer annotations'' that are linked to Wikidata entities and a few other value types.
This decision, we feel, restricts the range of natural language questions that are covered.
The cross-lingual aspect of the dataset appears to be primarily limited to entity translations, which likely do not cover a wide range of linguistic phenomena (which is the reason that we are interested in typologically diverse languages to begin with).
Thus, we believe that \mrtydi fills a gap in the evaluation space that is currently not occupied.

Many multi-lingual (both mono-lingual and cross-lingual) information retrieval and question answering datasets have been constructed over the past decades, via community-wide evaluations at TREC, FIRE, CLEF, and NCTIR.
These test collections are typically built on newswire articles, although some evaluations use Wikipedia and scientific texts.
While no doubt useful for evaluation, these test collections usually comprise only a small number of queries (at most a few dozen) with relevance judgments, which are insufficient to fine-tune dense retrieval models.
Furthermore, whereas \tydi at least draws from comparable corpora (i.e., Wikipedia articles), these test collections are built on corpora from much more diverse sources.
This makes it difficult to generalize across different languages.
For these reasons, the above-mentioned IR and QA test collections are not suitable for tackling the research questions we are interested in.

\section{\mrtydi}
\label{sec:dataset-prep}
\begin{table*}[!t]
\centering
\resizebox{1.5\columnwidth}{!}{
    \begin{tabular}{ll|rr|rr|rr|r}
    \toprule
    \multirow{2}{*}{} &
        & \multicolumn{2}{c}{\textbf{Train}} & \multicolumn{2}{c}{\textbf{Dev}} & \multicolumn{2}{c}{\textbf{Test}} & 
    \multirow{2}{*}{\textbf{Corpus Size}} \\
        & & \textbf{\# Q} & \textbf{\# J} & \textbf{\# Q} & \textbf{\# J} & \textbf{\# Q} & \textbf{\# J} & \\
    \midrule
    Arabic & (Ar) 
        & 12,377 & 12,377 & 3,115 & 3,115 & 1,081 & 1,257 & 2,106,586 \\
    Bengali & (Bn) 
        & 1,713 & 1,719 & 440 & 443 & 111 & 130 & 304,059 \\
    English & (En) 
        & 3,547 & 3,547 & 878 & 878 & 744 & 935 & 32,907,100 \\
    Finnish & (Fi)
         & 6,561 & 6,561 & 1,738 & 1,738 & 1,254 & 1,451 & 1,908,757 \\
    Indonesian & (Id)
        & 4,902 & 4,902 & 1,224 & 1,224 & 829 & 961 & 1,469,399 \\
    Japanese & (Ja) 
        & 3,697 & 3,697 & 928 & 928 & 720 & 923 & 7,000,027 \\
    Korean & (Ko) 
        & 1,295 & 1,317 & 303 & 307 & 421 & 492 & 1,496,126 \\
    Russian & (Ru) 
        & 5,366 & 5,366 & 1,375 & 1,375 & 995 & 1,168 & 9,597,504 \\
    Swahili & (Sw) 
        & 2,072 & 2,401 & 526 & 623 & 670 & 743 & 136,689 \\
    Telugu & (Te) 
        & 3,880 & 3,880 & 983 & 983 & 646 & 664 & 548,224 \\
    Thai & (Th)
        & 3,319 & 3,360 & 807 & 817 & 1,190 & 1,368 & 568,855 \\
    \midrule
    Total &
        & 48,729 & 49,127 & 12,317 & 12,431 & 8,661 & 10,092 & 58,043,326 \\
    \bottomrule
    \end{tabular}
}
\caption{Descriptive statistics for \mrtydi: the number of questions (\#~Q), judgments (\#~J), and the number of passages (Corpus Size) in each language.}
\label{tab:mr-tydi-stats}
\end{table*}

Having justified the need for a new benchmark dataset, this section describes the construction of \mrtydi, which can be best described as an open-retrieval extension to \tydi.

\paragraph{Corpus}
The formulation of any text ranking problem begins with a corpus $\mathcal{C} = \{ d_i \}$ comprising the units of text to be retrieved.
As the starting point, we used exactly the same raw Wikipedia dumps as \tydi.

Relevance annotations in \tydi are provided at the passage level (in the passage selection task), and thus we kept the same level of granularity in our corpus preparation.
For articles covered by \tydi (identified by the article titles), we retained the original passages.
For articles that are not covered by \tydi, we prepared passages using Wiki\-Extractor\footnote{\url{https://github.com/attardi/wikiextractor}} based on natural discourse units (e.g., two consecutive newlines in the wiki markup).
Unfortunately, \citet{clark-etal-2020-tydi} did not precisely document their passage segmentation method, but based on manual examination of our results, the generated passages appear qualitatively similar to the \tydi passages.

The result of the corpus preparation process is, for each language, a collection of passages from the Wikipedia articles in that language.
To form the final passages that comprise the basic unit of retrieval, we prepend the title of the Wikipedia article to each passage.
This creates retrieval units that can be more readily understood in isolation.\footnote{Mr.~\textsc{TyDi} v1.1 contains these article titles, whereas v1.0 did not. Results reported in this paper are with v1.1; for differences, please refer to the earlier version of our paper posted on arXiv.}

\paragraph{Task}
While \mrtydi is adapted from a QA dataset, our task is mono-lingual \textit{ad hoc} retrieval.
That is, given a question in language $L$, the task is to retrieve a ranked list of passages from $\mathcal{C}_L$, the Wikipedia collection in the same language (prepared in the manner described above), where the retrieved passages are ranked according to their relevance to the given question.

Our assumption here is a standard ``retriever--reader'' framework~\cite{chen-etal-2017-reading} or a multi-stage ranking architecture~\cite{Lin_etal_arXiv2020_ptr4tr}, where we focus on the retriever (what IR researchers call candidate generation or first-stage retrieval).
For end-to-end question answering, the output of the retriever would be fed to a reader for answer extraction.
This focus on retrieval allows us to explore the research questions outlined in the introduction, and this formulation is consistent with previous work in dense retrieval, e.g., \citet{karpukhin-etal-2020-dense}.

\paragraph{Questions and Judgments}
To prepare the questions, we started with all questions provided by \tydi and removed those without any answer passages or whose answer passages are all empty. 
We consider all non-empty annotated answer passages from \tydi as relevant to the corresponding question in \mrtydi. 
We adopt the development set of \tydi as our test set, since the original test data are not public.
A new development set was created by randomly sampling $20\%$ of questions from the original training set. 
We observed that some of the questions in \tydi are shared between the training and development set (but labeled with different answer passages).
In these cases, we retained the duplicate questions only in the training set. 
Descriptive statistics for \mrtydi are shown in Table~\ref{tab:mr-tydi-stats}, where languages are identified by their two letter ISO-639 language codes.

In summary, relevant passages in \mrtydi are imputed from \tydi.
Since \citet{clark-etal-2020-tydi} only asked annotators to assess the top-ranked article for each question, there are likely relevant passages that have not been identified.
Following standard assumptions in information retrieval, unjudged passages are considered non-relevant.
Thus, it is likely that ranking models will retrieve false negatives, i.e., passages that are relevant, but would not be properly rewarded.

In other words, our judgments are far from exhaustive.
This might be a cause for concern, but is a generally accepted practice in IR research due to the challenges of gathering complete judgments.
The widely used MS MARCO datasets~\cite{MS_MARCO_v3}, for example, share this characteristic of having ``sparse judgments''.
No claim is made about the exhaustiveness of the annotations, as both \mrtydi and MS MARCO provide only about one good answer per question.
From a methodological perspective, findings based on MS MARCO ``sparse judgments'' are largely consistent with results from more expensive evaluation efforts (to gather more complete judgments), such as the TREC Deep Learning Tracks~\cite{Craswell_etal_DL19_overview,Craswell_etal_DL20_overview}.
We expect a similar parallel here:\ more exhaustive judgments will change the absolute scores, but will likely not affect the findings qualitatively.

\paragraph{Metrics}
We evaluate results in terms of reciprocal rank and recall at a depth $k$ of 100 hits.
The first metric quantifies the ability of a model to generate a good ranking, while the second metric provides an upper bound on end-to-end effectiveness (e.g., when retrieval results are fed to a reader for answer extraction).
The setting of $k=100$ is consistent with work in the QA literature.

\section{Baselines}

We provide a few ``obvious'' baselines for \mrtydi as a starting point for future research:

\paragraph{BM25}
We report results with bag-of-words BM25, a strong traditional IR baseline, with the implementation provided by Pyserini~\cite{yang2017anserini,Lin_etal_SIGIR2021_Pyserini}, which is built on the open-source Lucene search library.
Lucene provides language-specific analyzers for nine of the eleven languages in \mrtydi; for these languages, we used the Lucene implementations. 
For Telugu (Te) and Swahili (Sw), since Lucene does not provide any language-specific implementations, we simply used its whitespace analyzer.
We report BM25 scores on two conditions, with default and tuned $k_1$ and $b$ parameters; the default settings are $k_1=0.9, b=0.4$.
Tuning was performed on the development set, on a per-language basis, via grid search on $k_1 \in [0.1, 1.6]$ and $b \in [0.1, 1.0]$, with step size $0.1$, optimizing MRR@100.

\paragraph{mDPR}
Dense passage retriever (DPR) by \citet{karpukhin-etal-2020-dense} is a well-known bi-encoder model for open-domain QA
%that separately encodes questions and snippets from Wikipedia using trained BERT encoders.
%Retrieval is performed using nearest neighbor search on the dense representations in terms of inner products.
%
that we adapt to mono-lingual retrieval in non-English languages by simply replacing BERT with multi-lingual BERT (mBERT),\footnote{Specifically, the \texttt{\small bert-base-multilingual-cased} model provided by HuggingFace~\cite{wolf-etal-2020-transformers}.}
but otherwise keeping all other aspects of the training procedure identical.
This adaptation, which we call mDPR, was trained on the English QA dataset Natural Questions~\cite{kwiatkowski2019natural} using Facebook's open-source codebase.

Our retrieval experiments with mDPR can be characterized as zero shot:\
We applied the same mBERT document encoder to convert passages from all eleven languages into dense vectors; similarly, we applied the same mBERT question encoder to all questions.
Retrieval in each language was performed using Facebook's Faiss library for nearest neighbor search~\cite{FAISS}; we used the FlatIP indexes.
Experiments were conducted using the same codebase as the DPR replication experiments of~\citet{ma2021replication}, with the Pyserini toolkit~\cite{Lin_etal_SIGIR2021_Pyserini} .

Our choice of zero-shot mDPR as a baseline deserves some discussion.
At a high level, we are interested in the generalizability of dense retrieval techniques in out-of-distribution settings (in this case, primarily different languages).
Operationally, our experimental setup captures the scenario where the model does not benefit from any exposure to the target task, even (question, relevant passage) pairs in the English portion of \mrtydi.
This makes the comparison ``fair'' to BM25, which is similarly not provided any labeled data from the target task (in the case with default parameters).

\begin{table*}[t]
\begin{subtable}{\textwidth}
\resizebox{\textwidth}{!}{
    \begin{tabular}{l|lllllllllll|l}
    \toprule
    & \textbf{Ar} & \textbf{Bn} & \textbf{En} & \textbf{Fi} & \textbf{Id} & \textbf{Ja} & \textbf{Ko} & \textbf{Ru} & \textbf{Sw} & \textbf{Te} & \textbf{Th} & \textbf{Avg} \\
    \midrule
    BM25 (default) & 
        0.368 & 0.418 & 0.140 & 0.284 & 0.376 & 0.211 & 0.285 & 0.313 & 0.389 & 0.343 & 0.401 & 0.321 \\
    BM25 (tuned) &
        0.367 & 0.413 & 0.151 & 0.288 & 0.382 & 0.217 & 0.281 & 0.329 & 0.396 & 0.424 & 0.417 & 0.333 \\
    mDPR & 
        0.260 & 0.258 & 0.162 & 0.113 & 0.146 & 0.181 & 0.219 & 0.185 & 0.073 & 0.106 & 0.135 & 0.167 \\
    hybrid & 
         0.491$^\dagger$ & 0.535$^\dagger$ & 0.284$^\dagger$ & 0.365$^\dagger$ & 0.455$^\dagger$ & 0.355$^\dagger$ & 0.362$^\dagger$ & 0.427$^\dagger$ & 0.405 & 0.420 & 0.492$^\dagger$ & 0.417 \\
    \bottomrule
    \end{tabular}
}
\caption{MRR@100}
\label{tab:results_mrr}
\end{subtable}

\vspace{1em}
\begin{subtable}{\textwidth}
\resizebox{\textwidth}{!}{
    \begin{tabular}{l|lllllllllll|l}
    \toprule
    & \textbf{Ar} & \textbf{Bn} & \textbf{En} & \textbf{Fi} & \textbf{Id} & \textbf{Ja} & \textbf{Ko} & \textbf{Ru} & \textbf{Sw} & \textbf{Te} & \textbf{Th} & \textbf{Avg} \\
    \midrule
    BM25 (default) & 
        0.793 & 0.869 & 0.537 & 0.719 & 0.843 & 0.645 & 0.619 & 0.648 & 0.764 & 0.758 & 0.853 & 0.732 \\
    BM25 (tuned) & 
        0.800 & 0.874 & 0.551 & 0.725 & 0.846 & 0.656 & 0.797 & 0.660 & 0.764 & 0.813 & 0.853 & 0.758 \\
    mDPR & 
        0.620 & 0.671 & 0.475 & 0.375 & 0.466 & 0.535 & 0.490 & 0.498 & 0.264 & 0.352 & 0.455 & 0.473 \\
    hybrid & 
        0.863$^\dagger$ & 0.937 & 0.696$^\dagger$ & 0.788$^\dagger$ & 0.887$^\dagger$ & 0.778$^\dagger$ & 0.706$^\dagger$ & 0.760$^\dagger$ & 0.786 & 0.827 & 0.875$^\dagger$ & 0.809 \\
    \bottomrule
    \end{tabular}
}
\caption{Recall@100}
\label{tab:results_recall1k}
\end{subtable}

\caption{Results of BM25 (with default and tuned parameters), mDPR, and the sparse--dense hybrid on the test set of \mrtydi.
The symbol $^\dagger$ indicates significant improvements over BM25 (tuned) (paired $t$-test, $p < 0.01$).}
\label{tab:results}
\end{table*}

\paragraph{Sparse--Dense Hybrid}
Our hybrid technique combines the scores of sparse (BM25) and dense (mDPR) retrieval results.
The final fusion score of each document is calculated by $s_{\small \textrm{sparse}} + \alpha \cdot s_{\small \textrm{dense}}$,
where $s_{\small \textrm{sparse}}$ and $s_{\small \textrm{dense}}$ represent the scores from sparse and dense retrieval, respectively.
This strategy is similar to the one described by \citet{ma2021replication}.
We take 1000 hits from mDPR and 1000 hits from BM25 and normalize the scores from each into $[0,1]$ since the range of the two types of scores otherwise are quite different.
If one hit isn't found in the other, the normalized score for that hit is set to zero.
The weight $\alpha$ was tuned in $[0,1]$ with a simple line search on the development set by optimizing MRR@100 with step size 0.01.

\section{Results and Analysis}

We performed experiments on \mrtydi v1.1, where each passage contains the title of the Wikipedia article and the passage text.
Table~\ref{tab:results} reports results on the test set across all eleven languages; mean reciprocal rank (MRR) in the top table and recall in the bottom table, both at a cutoff of 100 hits; the final column reports the average across all languages.
The rows report BM25 results (default and tuned), followed by results of mDPR and the sparse--dense hybrid.
For the hybrid method, statistically significant improvements over tuned BM25 are denoted with the symbol $^\dagger$ based on paired $t$-tests ($p < 0.01$).

\begin{figure*}[!t]
\centering
\includegraphics[trim={0 0 0 3em},clip, width=0.9\columnwidth]{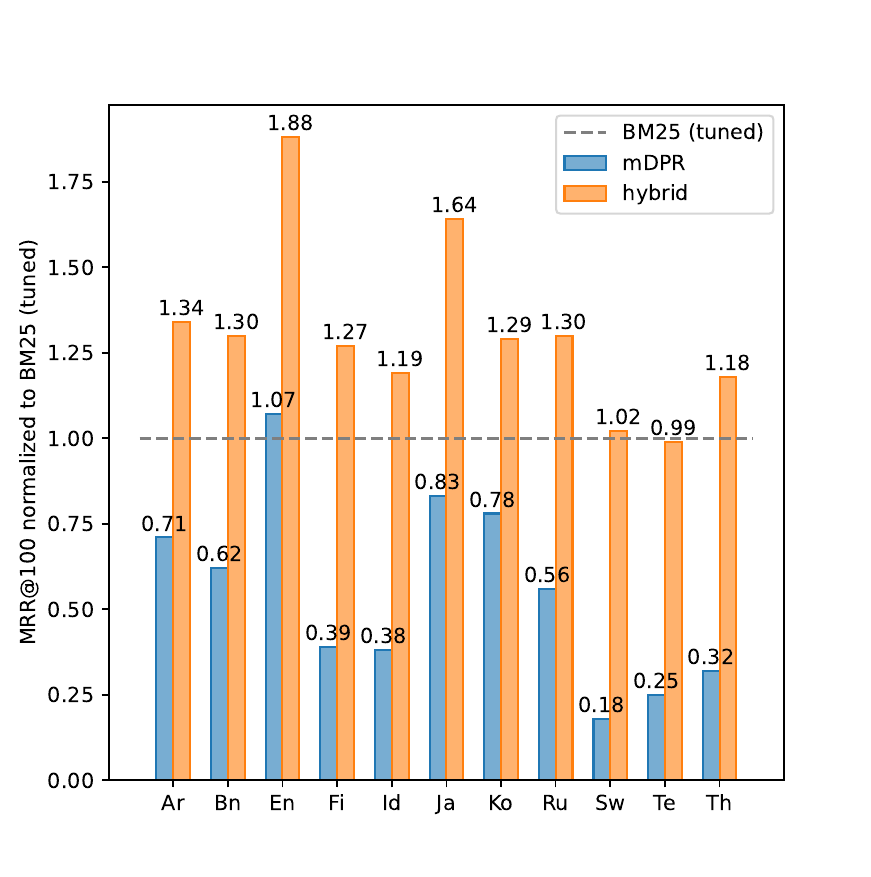}
\includegraphics[trim={0 0 0 3em},clip,width=0.9\columnwidth]{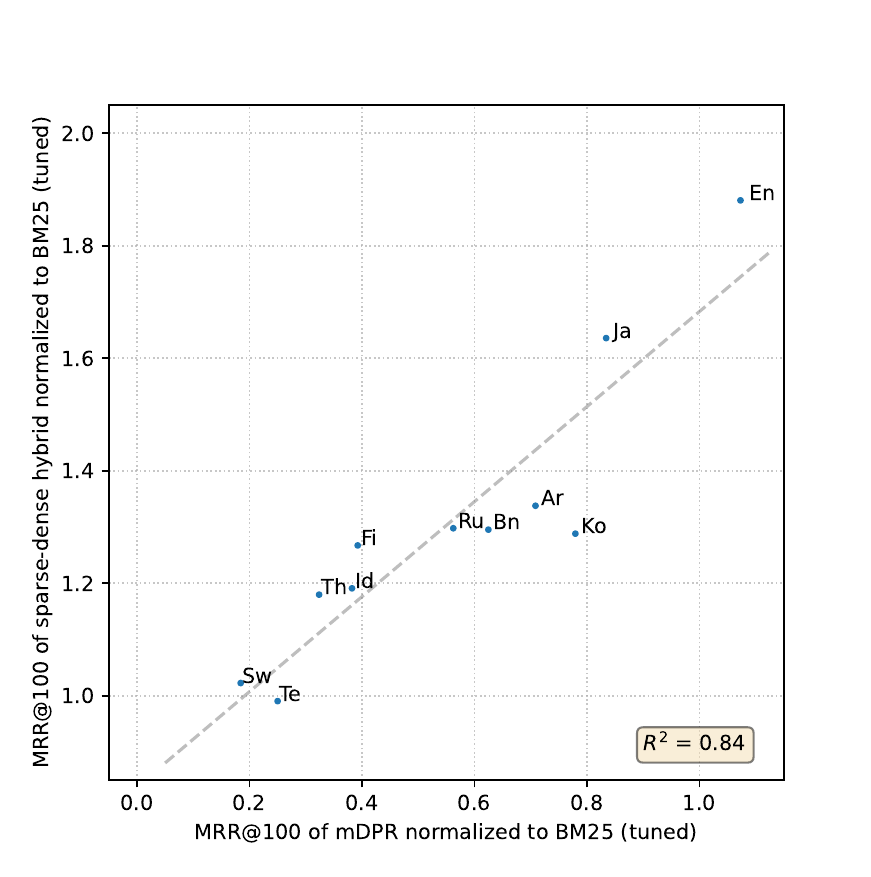}
\caption{MRR@100 of mDPR and the sparse--dense hybrid normalized with respect to BM25 for each language (left); corresponding pairs for each language plotted as a scatter plot (right).}
\label{fig:relative-scores}
\end{figure*}

\begin{figure*}[p]
\centering
      \includegraphics[width=.24\textwidth]{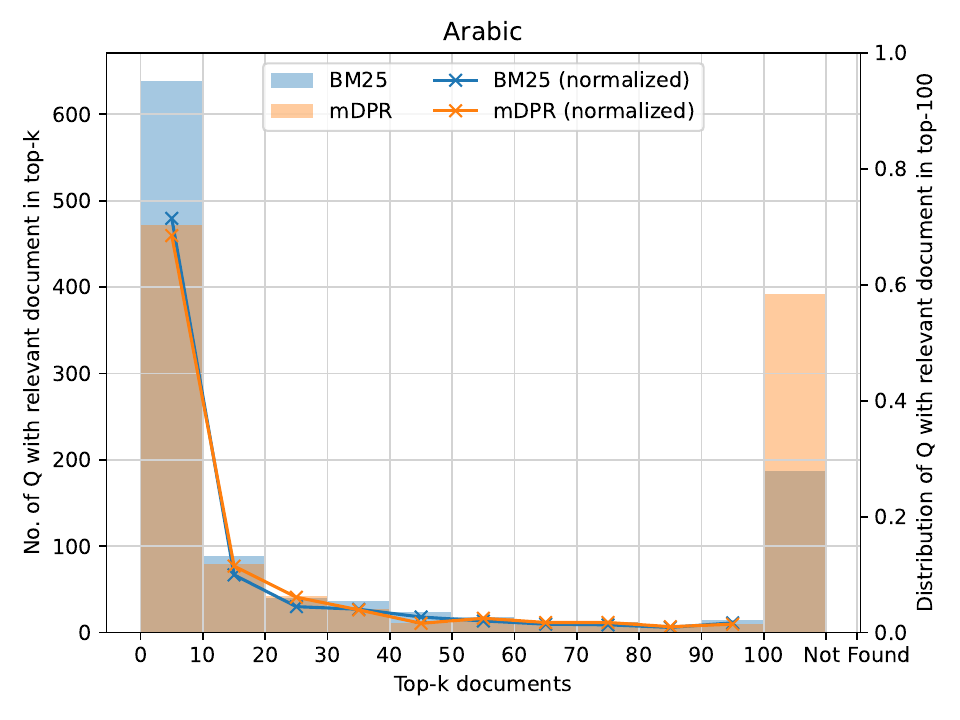}
      \includegraphics[width=.24\textwidth]{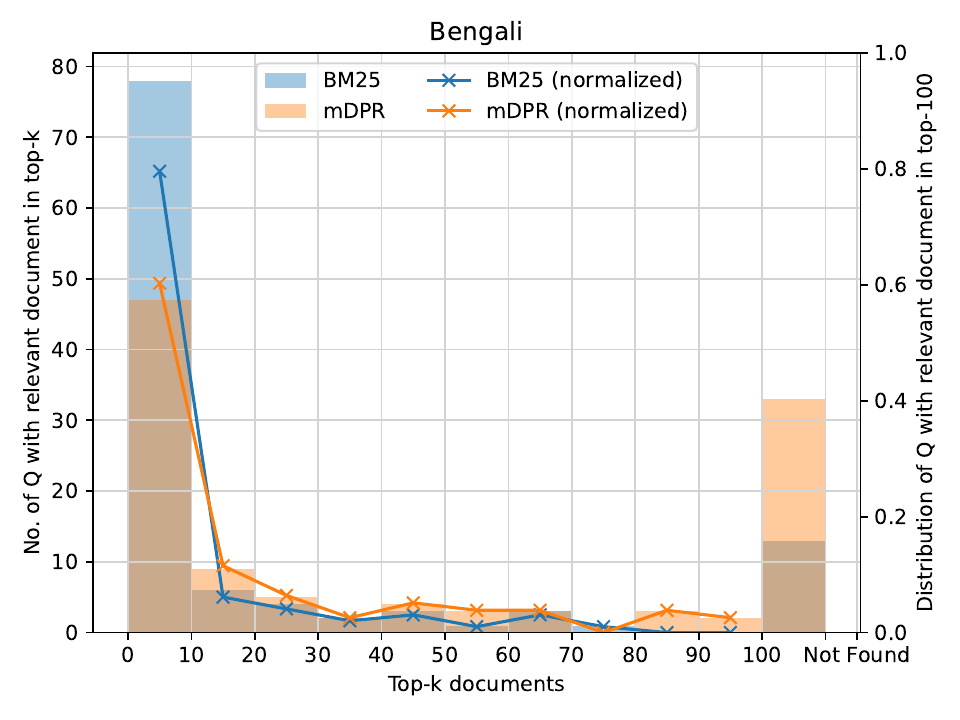}
      \includegraphics[width=.24\textwidth]{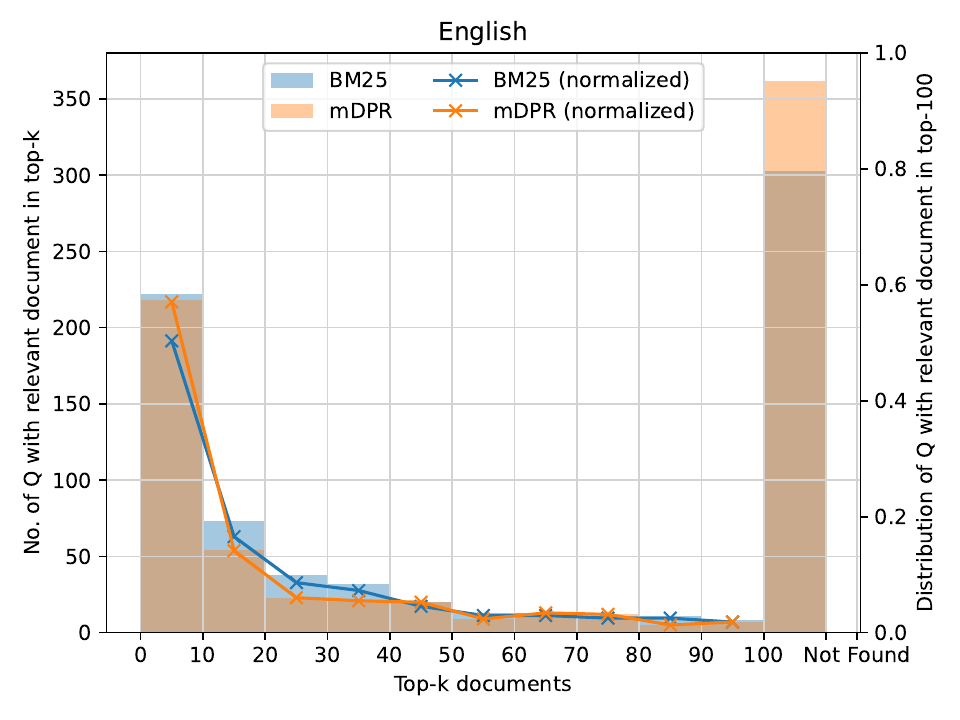}
      \includegraphics[width=.24\textwidth]{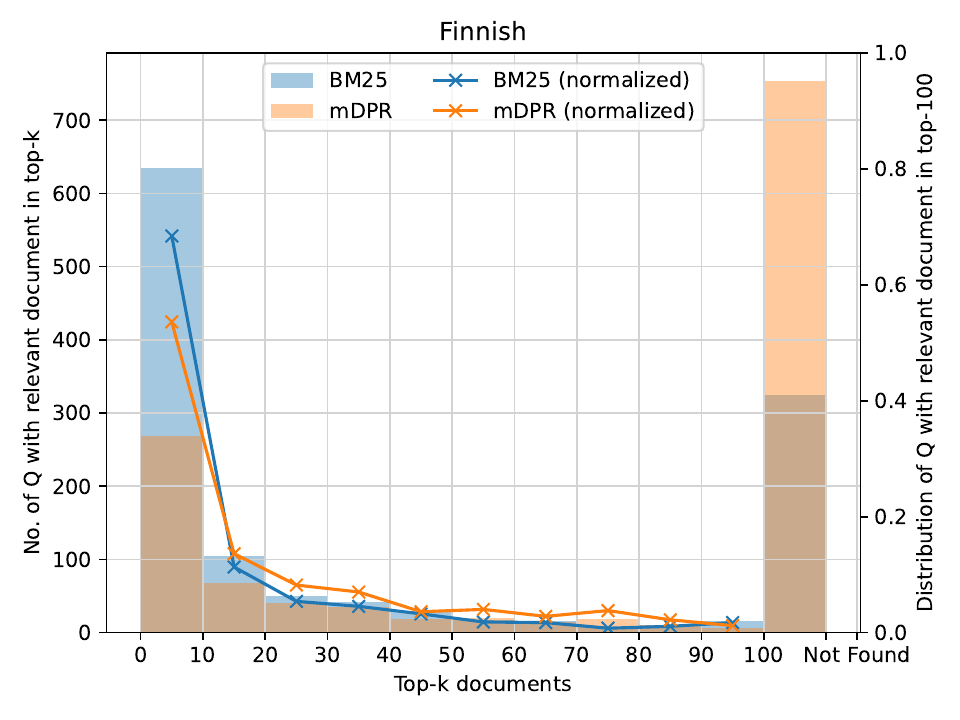}
      \includegraphics[width=.24\textwidth]{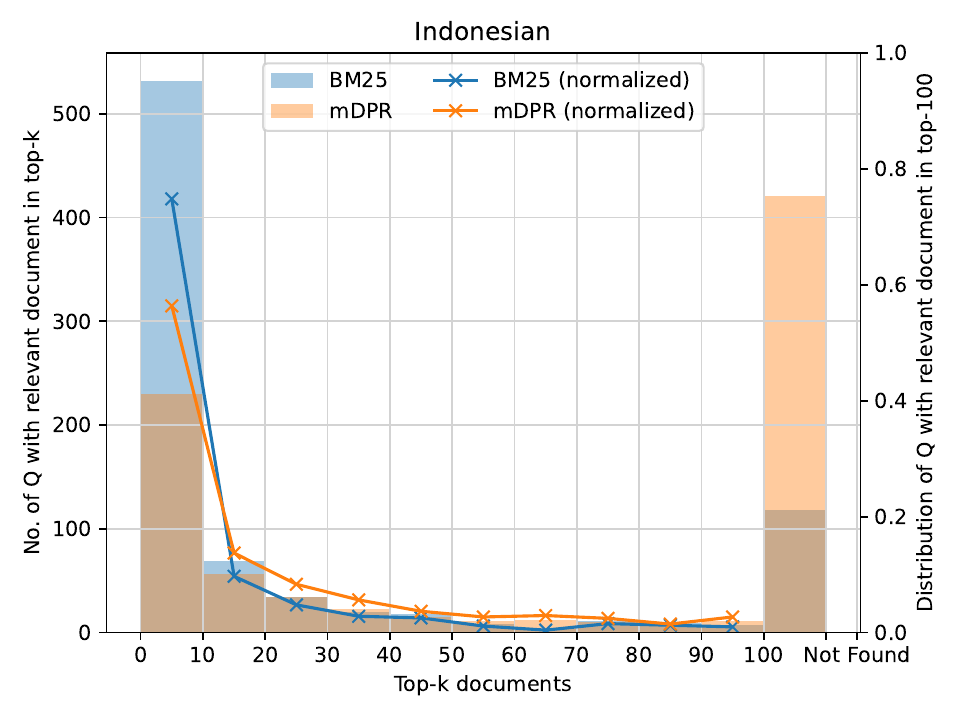}
      \includegraphics[width=.24\textwidth]{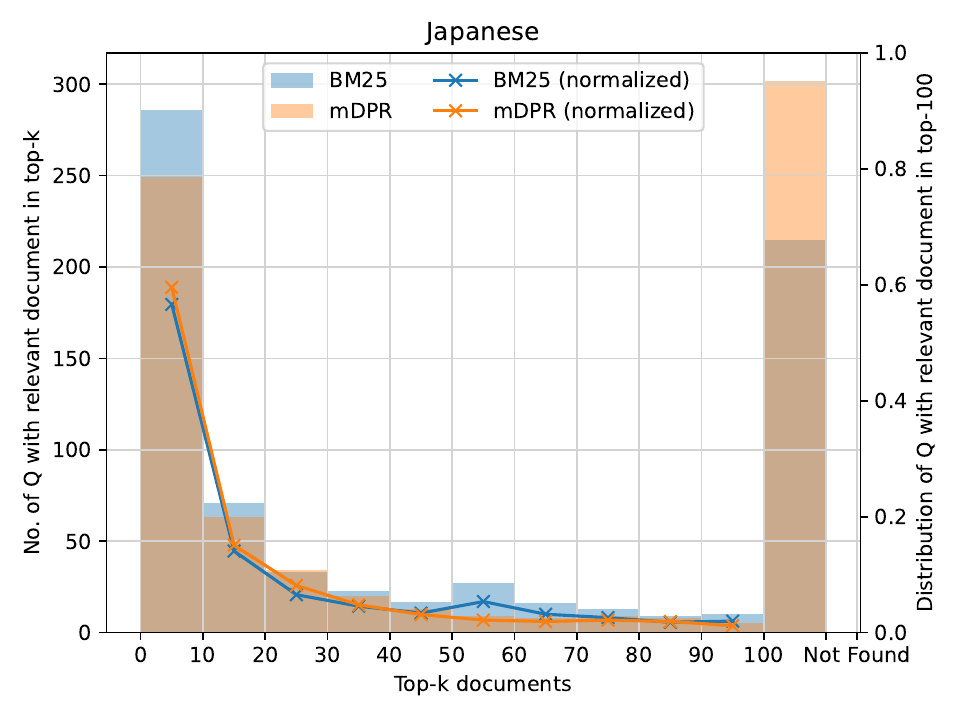}
      \includegraphics[width=.24\textwidth]{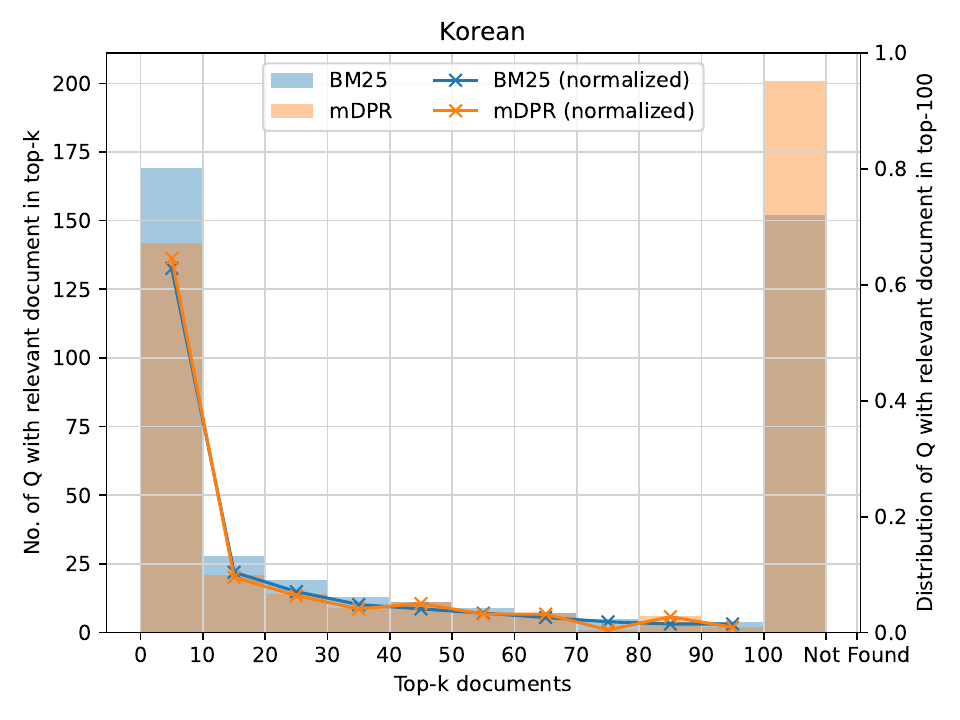}
      \includegraphics[width=.24\textwidth]{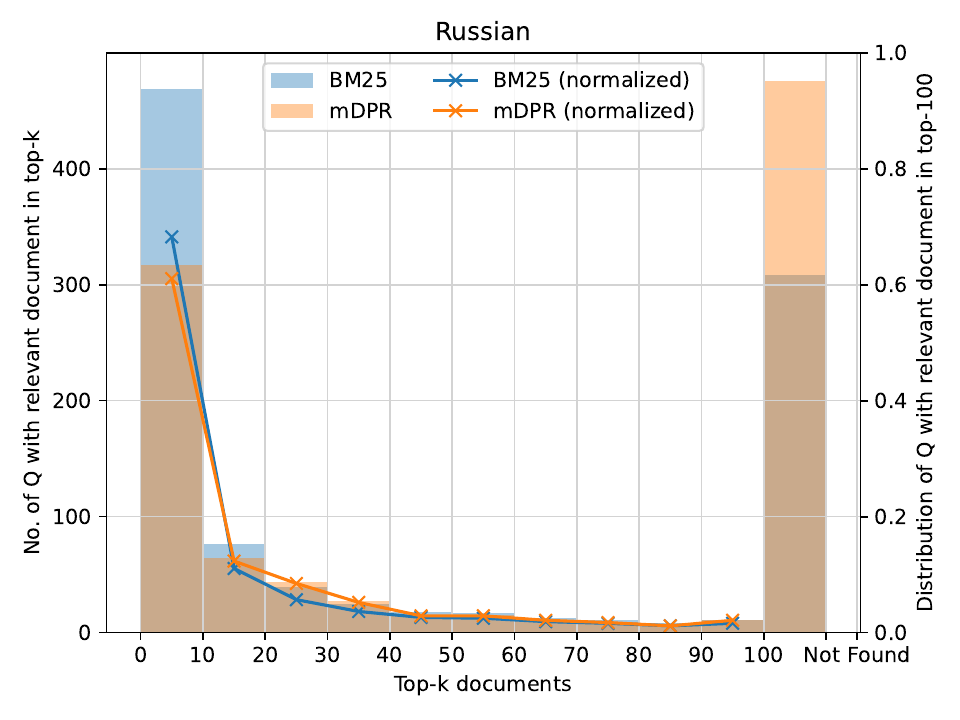}
      \includegraphics[width=.24\textwidth]{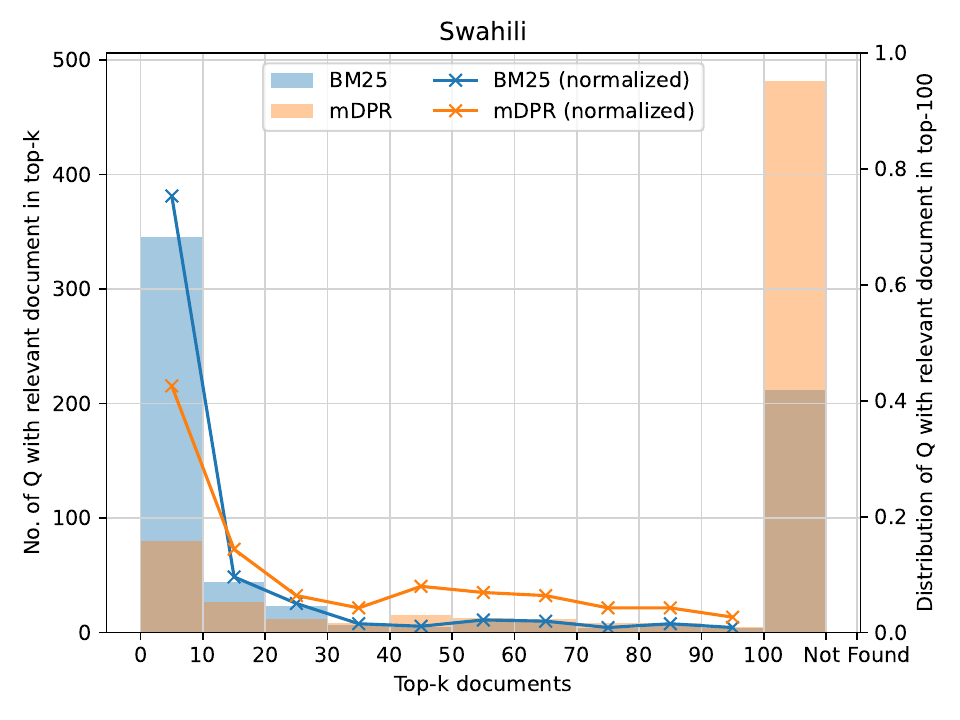}  
      \includegraphics[width=.24\textwidth]{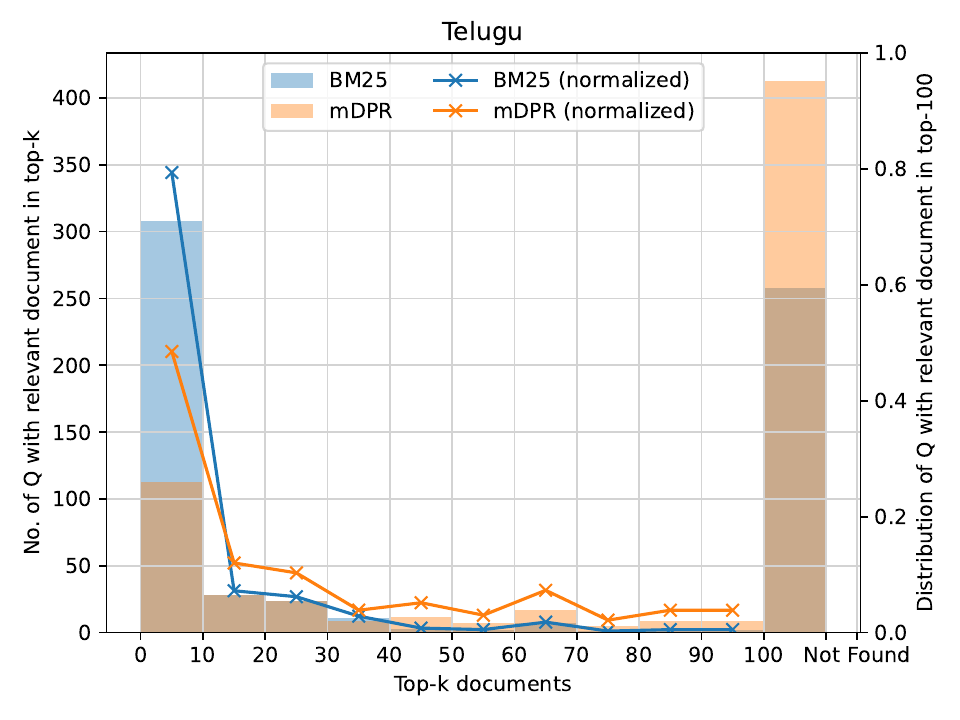}
      \includegraphics[width=.24\textwidth]{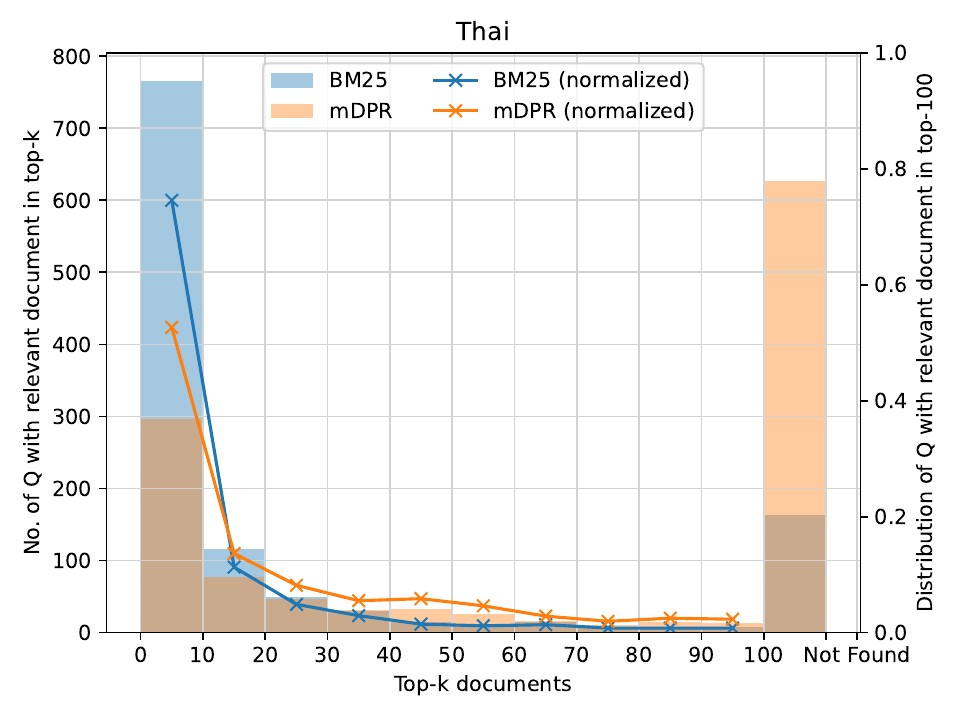}
    % \end{subfigure}
\caption{Analysis of recall and ranking effectiveness comparing BM25 (tuned) and mDPR. In each plot, the histogram shows the distribution of relevant passages; lines plot the distribution of relevant passages normalized to only questions where a relevant passage appears in the top-100 hits.}
\label{fig:analysis-dense}
\end{figure*}

\begin{figure*}[p]
\centering
      \includegraphics[width=.24\textwidth]{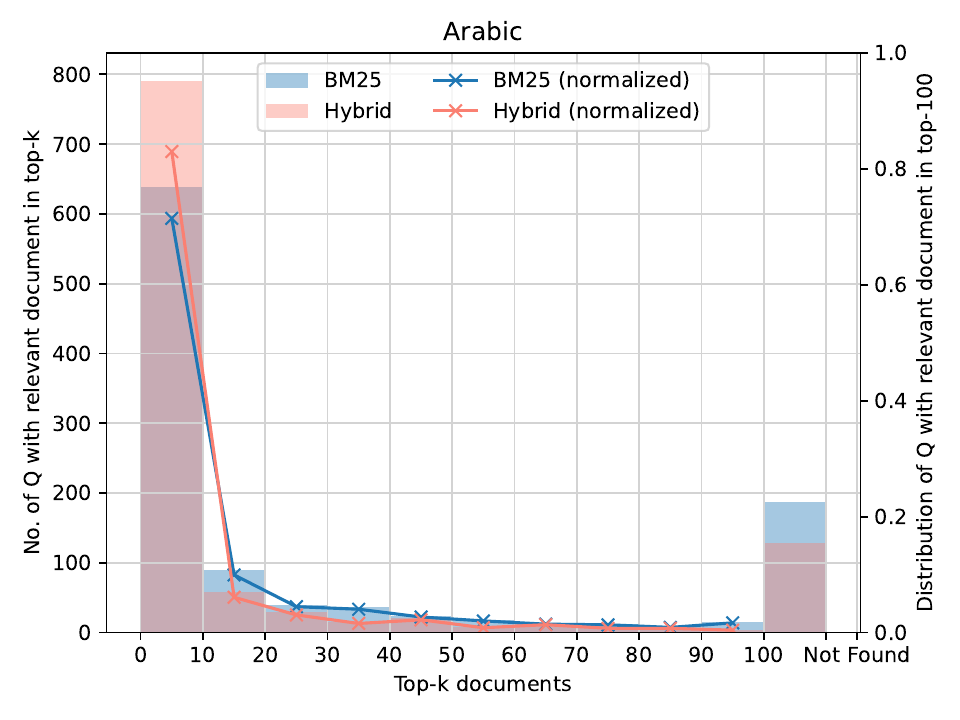}
      \includegraphics[width=.24\textwidth]{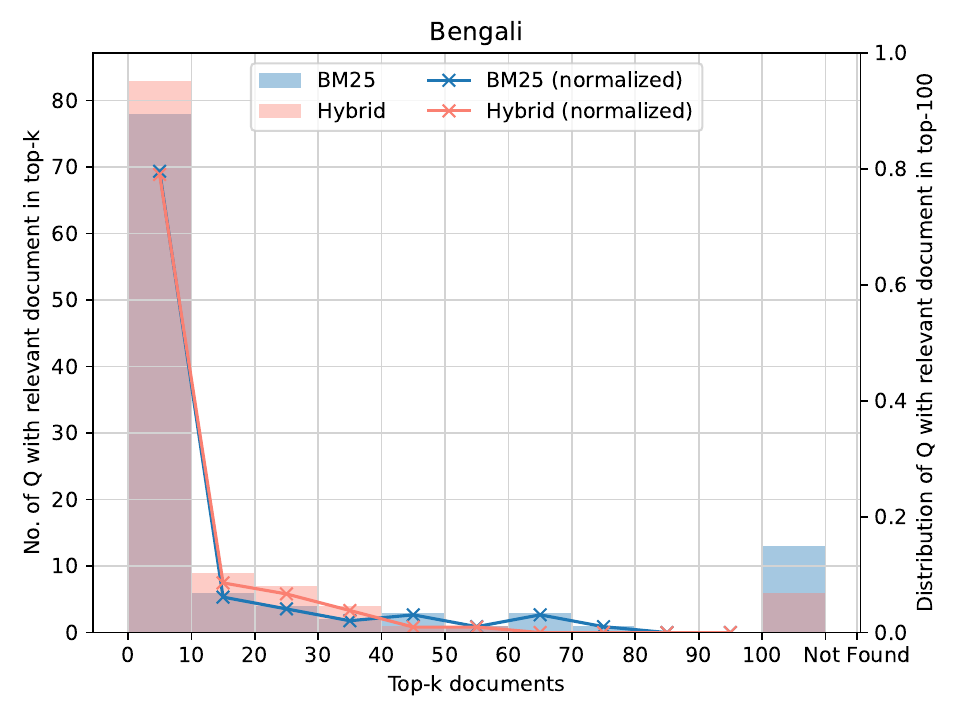}
      \includegraphics[width=.24\textwidth]{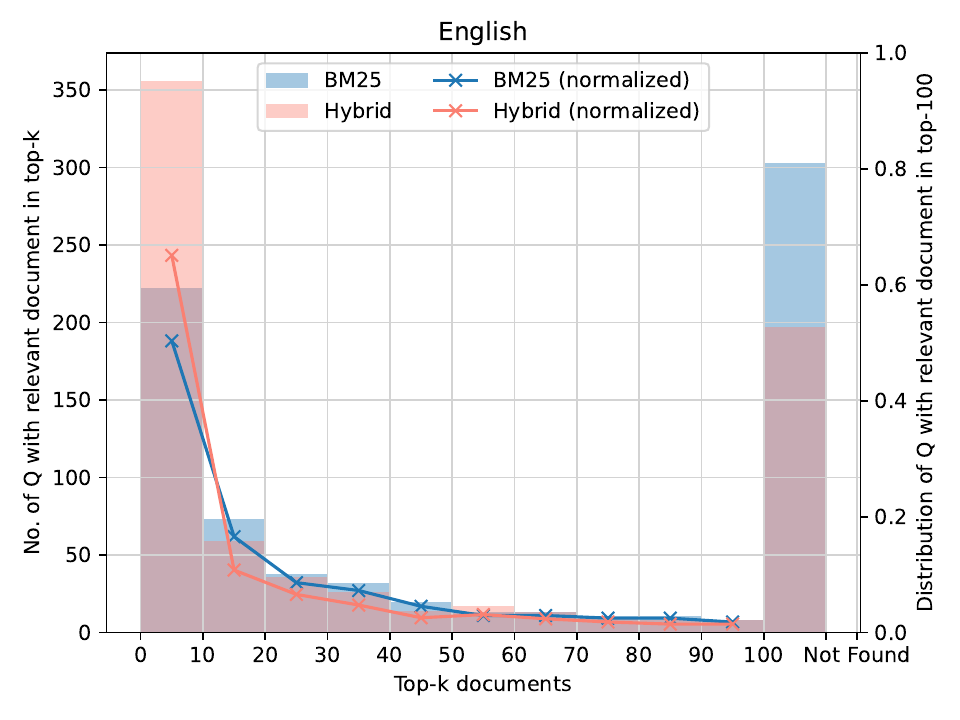}
      \includegraphics[width=.24\textwidth]{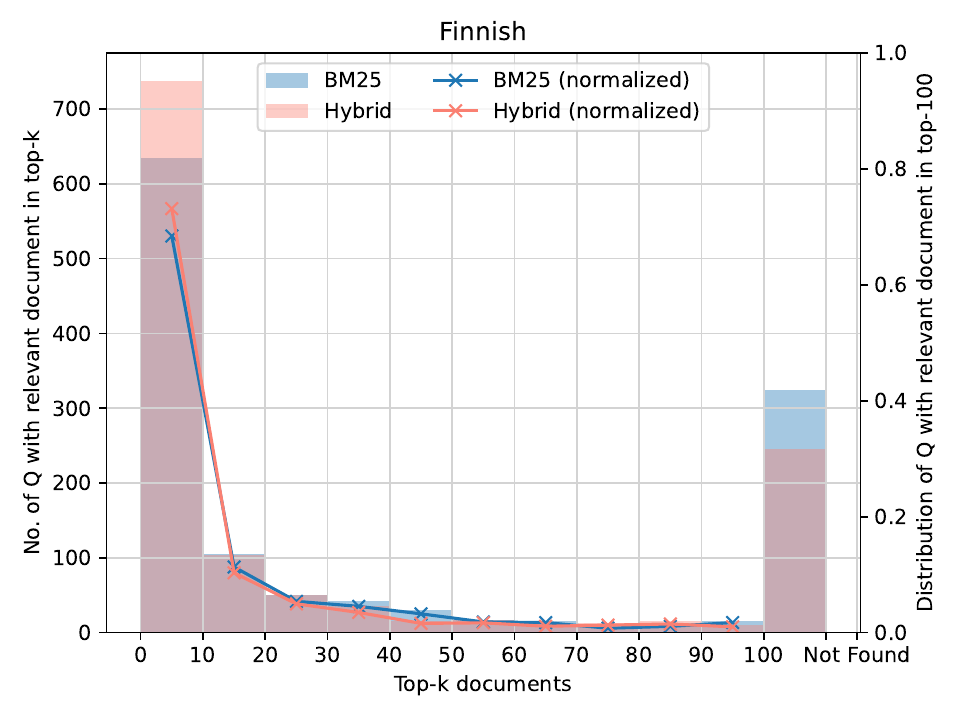}
      \includegraphics[width=.24\textwidth]{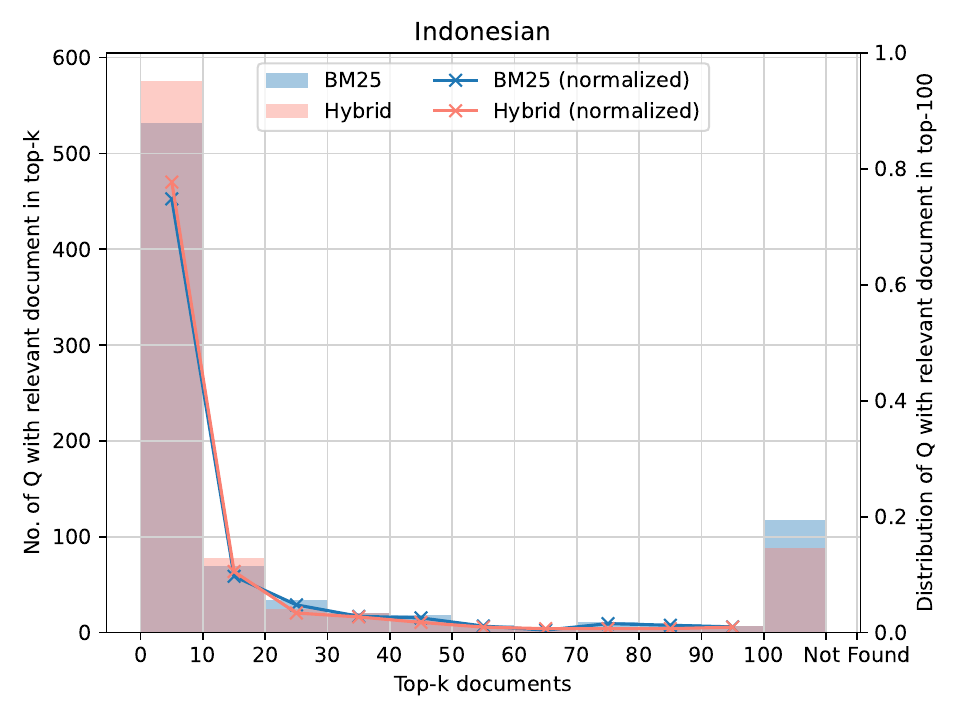}
      \includegraphics[width=.24\textwidth]{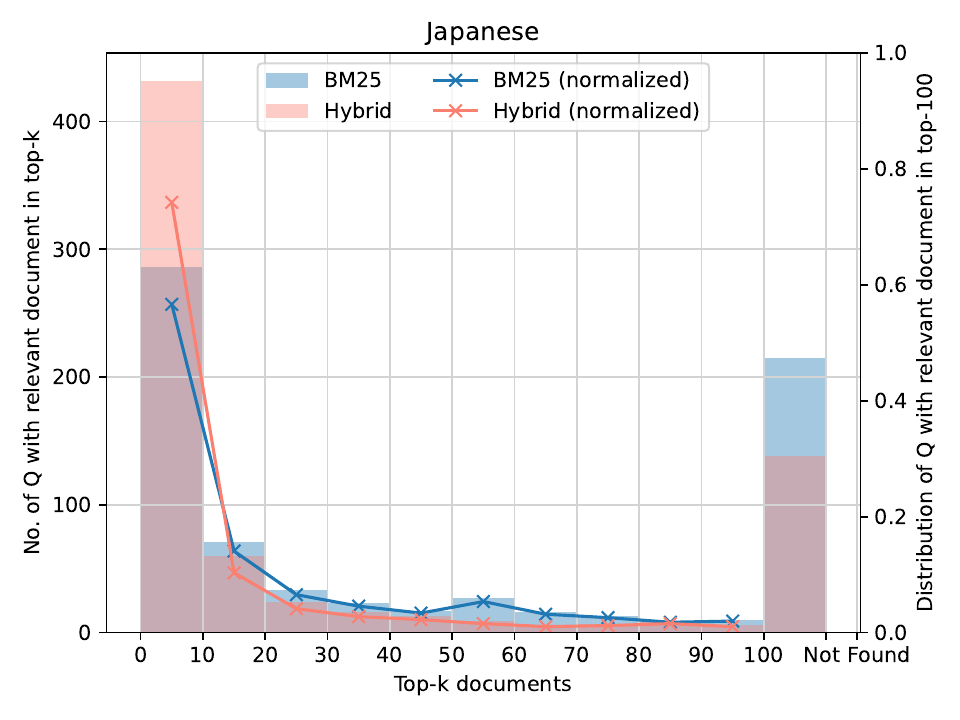}
      \includegraphics[width=.24\textwidth]{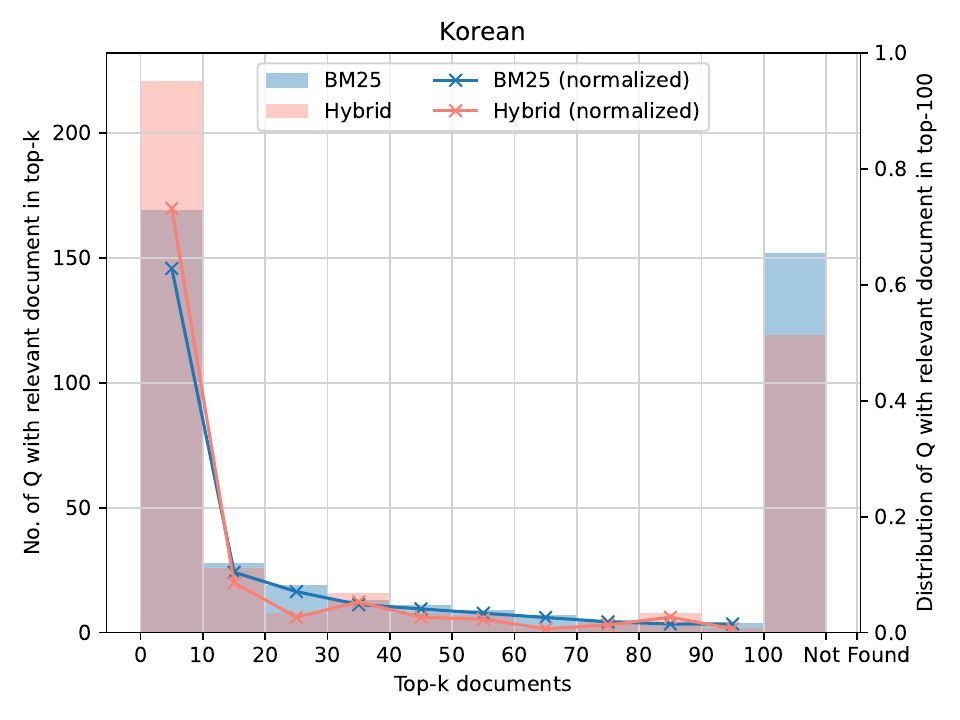}
      \includegraphics[width=.24\textwidth]{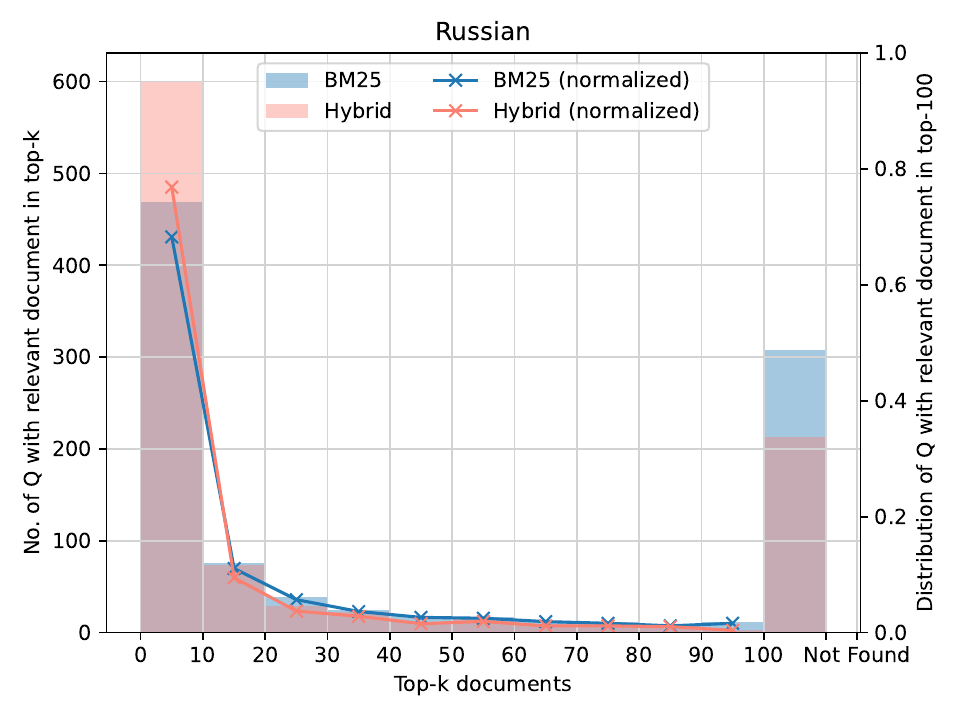}
      \includegraphics[width=.24\textwidth]{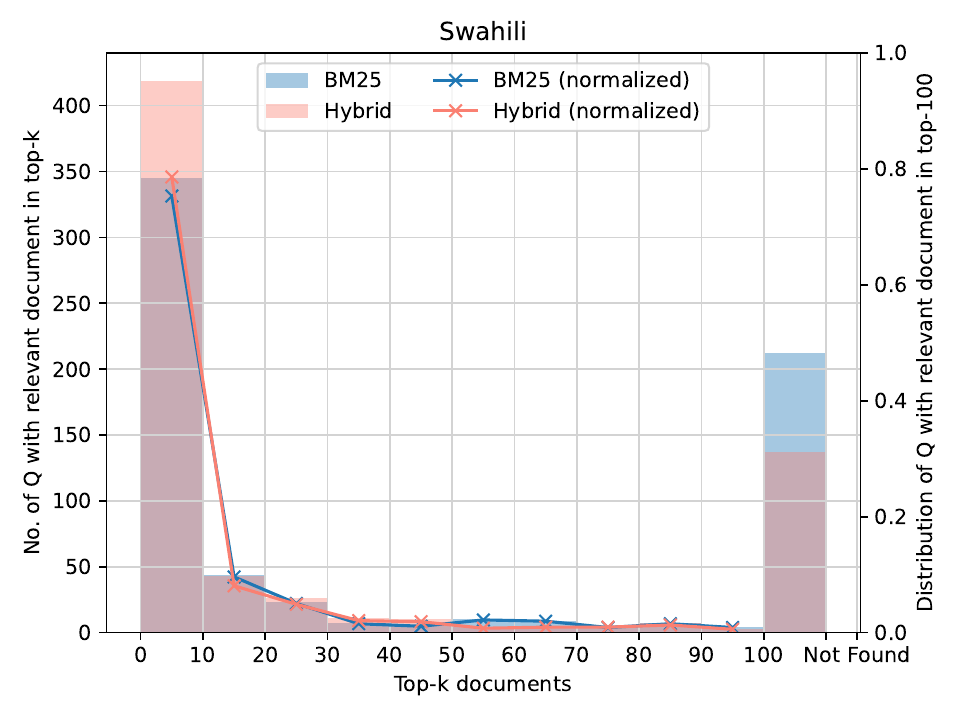}
      \includegraphics[width=.24\textwidth]{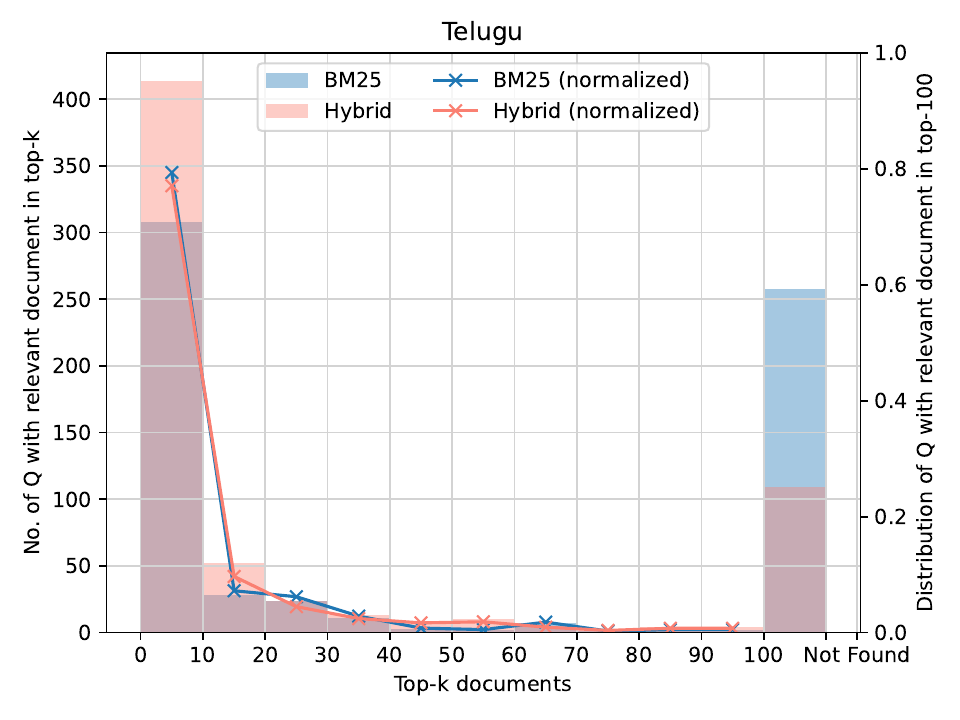}
      \includegraphics[width=.24\textwidth]{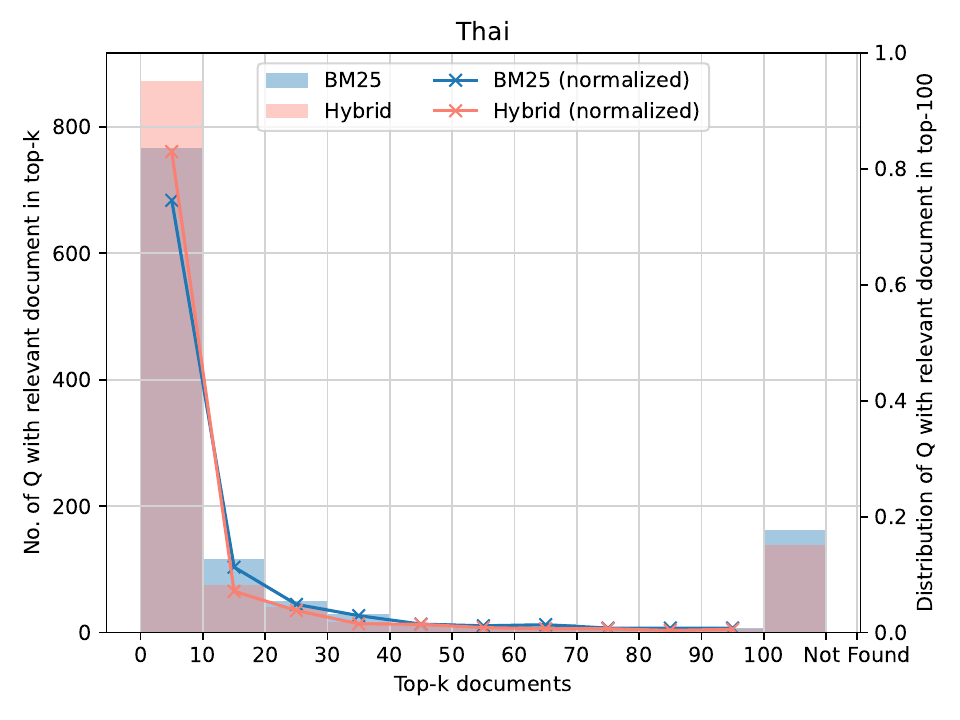}

\caption{Analysis of recall and ranking effectiveness comparing BM25 (tuned) and the sparse--dense hybrid. Each plot is constructed in a similar manner as the plots in Figure~\ref{fig:analysis-dense}.}
\label{fig:analysis-hybrid}
\end{figure*}

\subsection{High-Level Findings}
\label{sec:findings}

By comparing scores in each column, we observe that the absolute effectiveness of the techniques varies greatly across languages.
Absolute scores are difficult to compare because both the questions and the underlying corpora are different.
However, three high-level findings emerge:

First, we find that tuning BM25 parameters yields at most minor improvements for most languages, both in terms of MRR@100 and recall except for Telugu (cases where scores decrease slightly can be explained by noise in the training/test splits).
This is a bit of a surprise, as parameter tuning usually yields larger overall gains, e.g., in the MS MARCO collections~\cite{MS_MARCO_v3}.
Regardless, tuned BM25 serves as a competitive baseline for the remainder of our experiments.

Second, we notice that mDPR underperforms BM25 across all languages except for English.
That is, in a zero-shot setting, retrieval using learned dense representations from mDPR (fine-tuned with NQ) is a lot worse than just retrieval using BM25-based representations.
Clearly, mDPR is far less robust in cross-lingual generalizations.
Even within the same language, mDPR seems to be sensitive to characteristics of the training data.
Effectiveness on the English portion of \mrtydi is only slightly better than BM25, likely arising from the fact that we are applying an NQ-trained model on ``out-of-distribution'' questions.

Based on~\citet{karpukhin-etal-2020-dense} and the experiments by~\citet{ma2021replication}, we would have expected mDPR to beat BM25 for in-distribution training and inference.
Since NQ is also based on Wikipedia, corpus differences are less likely an issue; these results suggest that questions in \tydi and NQ are qualitatively different.

Third, despite the fact that mDPR effectiveness is quite a bit worse than BM25, the MRR@100 of the sparse--dense hybrid is significantly higher than tuned BM25 for nine of the eleven languages (the exceptions are Swahili and Telugu).
Rephrased differently, this means that although mDPR by itself is a poor dense retrieval model in a zero-shot setting, it nevertheless contributes valuable relevance signals that are able to improve over tuned BM25.
On average, the hybrid results are around eight and five points absolute higher than tuned BM25 in terms of MRR@100 and recall, respectively.

Because absolute scores vary widely across languages, it is helpful to normalize the effectiveness of tuned BM25 to 1.0 and scale the effectiveness of mDPR and the hybrid approach appropriately; this is shown in Figure~\ref{fig:relative-scores} (left) for MRR@100.
As an example, from the leftmost bars, we see that the MRR@100 of mDPR in Arabic is 71\% of BM25 but the sparse--dense hybrid improves over BM25 by 34\% (stat.\ sig).
We additionally plot the relation between the normalized effectiveness of mDPR and the hybrid approach in Figure~\ref{fig:relative-scores} (right).

This plot shows a clear positive (linear) correlation, that is, better mDPR (relative) effectiveness translates into bigger improvements over BM25 in the sparse--dense hybrid.
What is surprising, though, is that this relationship seems to hold even if the mDPR results are poor.
For example, in Thai, the MRR@100 of mDPR is only 32\% of BM25 (tuned), yet the hybrid yields a statistically significant 18\% relative gain in the hybrid approach.
However, there appear to be limits to our simple linear combination of relevance signals:\ for both Swahili and Telugu, the hybrid approach does not outperform tuned BM25, likely because mDPR effectiveness is too poor.

\subsection{Components of Effectiveness}

To provide a more in-depth analysis, we attempt to untangle effectiveness into two separate components:\ (1) retrieving a relevant passage and (2) placing the relevant passages into top ranks.
The recall figures in Table~\ref{tab:results} already quantify the first component, but MRR@100 alone does not tell the complete story for the second component, since the metric averages a bunch of zeros for questions where relevant passages do not appear in the top-100 hits.
It could be the case, for example, that mDPR provides a good ranking for those queries where it retrieves a relevant result in the top 100.

The results of such an analysis, comparing BM25 (tuned) and mDPR, are shown in Figure~\ref{fig:analysis-dense} for all languages (ordered alphabetically).
Each plot consists of a histogram and a line graph.
The histogram captures the distribution of the ranks (binned by ten) where the relevant passage appears for each question.\footnote{If the question has multiple retrieved relevant passages, we only consider the smallest rank among them (i.e., the highest ranked relevant passage).}
Questions for which no relevant passage was found in the top-100 hits are tallied in the rightmost bar (``Not Found'').
Thus, all questions are either in the rightmost bar (not found in the top-100 hits) or in one of the top-100 bins; these are exactly the components of recall, so the histograms are a more fine-grained way to visualize recall.

The superimposed line graphs in each plot show the ratio of the number of questions falling in each bin to the total number of questions in all top-100 bins (that is, we remove the ``Not Found'' bin and renormalize).
These plots answer the following question:\ Given that the relevant passage appeared in the top-100 hits, how well did the model perform at ranking it?
In other words, we have isolated the ranking ability of the model.

Looking only at the line graphs, these results tell us that for Arabic, Japanese, and Korean, BM25 and mDPR are comparable when we focus only on ranking---that is, given that the relevant passage appears in the top-100 hits.
In other words, MRR@100 differences for these languages come mostly from the fact that mDPR misses many relevant passages that BM25 finds (i.e., exhibits lower recall).
For the other languages, BM25 appears to exhibit {\it both} better recall and better ranking.
Consider Swahili, for example, BM25 places many more relevant passages in the top 10 and also has far fewer questions where no correct answer appears in the top-100 hits.
Thus, this analysis isolates the different failure modes of mDPR (dense retrieval) relative to BM25 (sparse retrieval).

The same analysis comparing BM25 (tuned) and the sparse--dense hybrid is shown in Figure~\ref{fig:analysis-hybrid}.
These plots reveal how the hybrid is improving the BM25 results.
We see that gains in Bengali, Indonesian, Swahili, and Telugu come mostly from higher recall.
That is, ranking capabilities are roughly comparable (the line plots largely overlap) but the hybrid approach has fewer queries where the relevant passage does not appear in the top-100 hits.
For Thai, the gain comes from better ranking, while recall is just a small bit better (the ``Not Found'' bars are pretty close).
For the other languages, hybrid improves both recall and ranking.

\section{Future Work}

\mrtydi provides a resource to begin exploring mono-lingual {\it ad hoc} retrieval with both dense and sparse retrieval techniques.
In this paper, we have focused primarily on zero-shot baselines.
Although zero-shot dense retrieval (mDPR) does not appear to be effective by itself, relevance signals from the model do appear to be complementary to sparse retrieval (bag-of-words BM25).
We have identified {\it how} they are complementary (better recall vs.\ better ranking), but the behavior varies across languages, and we do not yet have an explanation for {\it why}; for example, do the typological characteristics of the language play a role?

For our experiments, we have decided to focus on zero-shot effectiveness because it serves as the natural baseline of any technique that tries more sophisticated approaches.
Thus, the baselines here are foundational to any future work.
We have explicitly decided not to report any language-specific fine-tuning results here, although preliminary experiments suggest that such techniques do bring about benefits.
We have not yet systematically explored the broad design space of what~\citet{Lin_etal_arXiv2020_ptr4tr} calls ``multi-step fine-tuning strategies'', paralleling the explorations of~\citet{Shi_etal_FindingsEMNLP2020} in the context of transformer-based reranking models.
There are many possible variations, for example, how many languages to use, what order to sequence data from different languages, possible data augmentation using machine translation, complementary data from other tasks, etc.
There are a number of experiments that will allow us to tease apart the effects of language versus other aspects of training data distribution (e.g., NQ vs.\ \tydi).
Exploration of this vast design space is the focus of our immediate future work.

In addition, we believe that our dataset can provide a probe to examine the nature of multi-lingual transformer models.
Our experimental results show that absolute effectiveness varies quite a bit across languages.
Some of these variations may be due to the nature of the queries, the size of the corpora, etc.
However, we hypothesize that inherent properties of the transformer model play important roles as well, e.g., the size of the pretraining corpus in each language, typological and other innate characteristics of the languages, etc.
We hope that \mrtydi can help us untangle some of these issues.

\section{Conclusion}

In this work, we introduce \mrtydi, a multi-lingual benchmark dataset for mono-lingual retrieval in eleven typologically diverse languages, built on \tydi.
We describe zero-shot experiments using BM25, mDPR, and a sparse--dense hybrid.
The experimental results are not surprising:\ as is already known from complementary experiments, dense retrieval techniques do not generalize well to out-of-distribution input.
However, we find that even poor dense retrieval results provide valuable relevance signals in a sparse--dense hybrid.

Of course, this is only the starting point.
With \mrtydi, we now have a resource to explore our motivating research questions regarding the behavior of dense retrieval models when fed ``out of distribution'' data, and from there, devise techniques to increase the robustness and generalizability of our techniques.
The potential broader impact of this work is more equitable distribution of information access capabilities across diverse languages of the world---to help non-English speakers access relevant information in their own languages.

\section*{Acknowledgements}

This research was supported in part by the Canada First Research Excellence Fund and the Natural Sciences and Engineering Research Council (NSERC) of Canada; computational resources were provided by Compute Ontario and Compute Canada.

\bibliography{anthology}
\bibliographystyle{acl_natbib}

\end{document}